\documentclass[conference,letterpaper]{IEEEtran}
\usepackage{fancyhdr,graphicx}
\begin{document}

\title{Microcontroller-based System for Modular Networked Robot}

\author{\authorblockN{I. Firmansyah, Z. Akbar, B. Hermanto and L.T. Handoko}
\authorblockA{Group for Theoretical and Computational Physics, Reearch Center for Physics, Indonesian Institute of Sciences (LIPI)\\
Kompleks Puspiptek Serpong, Tangerang 15310, Indonesia\\
firmansyah@teori.fisika.lipi.go.id, zaenal@teori.fisika.lipi.go.id}}

\maketitle

\begin{abstract}
A prototype of modular networked robot for autonomous monitoring works with full control over web through wireless connection has been developed. The robot is equipped with a particular set of built-in analyzing tools and appropriate censors, depending on its main purposes, to enable self-independent and real-time data acquisition and processing. The paper is focused on the microcontroller-based system to realize the modularity. The whole system is divided into three modules : main unit, data acquisition and data processing, while the analyzed results and all aspects of control and monitoring systems are fully accessible over an integrated web-interface. This concept leads to some unique features : enhancing flexibility due to enabling  partial replacement of the modules according to  user needs, easy access over web for remote users, and low  development and maintenance cost due to software dominated components. 
\end{abstract}

\IEEEpeerreviewmaketitle

\section{Introduction}

Nowadays, automated systems embedded with remote robots are getting common in both manufacturing and non-manufacturing industries. In non-manufacturing sectors, it is motivated in most cases by the concern of safety as volcano observations and so forth. Following recent internet technologies, many groups have developed the so-called networked robots that is robotic systems controlled remotely over internet  using TCP/IP protocol. Such technologies are mainly applied to support human daily life, or realize more interactive humanoids. One example is the WAX Project which is the second tele-operated internet robot at Ryerson Polytechnic Universities, the MAX Project \cite{max3}. The MAX Tele-Operated Dog has shown that  a tele-operated robot controlled from over web is quite reliable \cite{max1}\cite{max2}. On the other hand, the WAX puts together a procedure to change any robot into a tele-operated robot on the web \cite{wax}. Either MAX / WAX or MONEA are the microcontroller-based robots equipped with onboard computer, camera and  microphone with a main purpose to simulate telepresence. Originally these robots were intended and  simulated to support the handicapped persons. So the main issue is how to recognize the captured images or sounds and interpret them to be useful information for potential users.

Another kind of networked robot is the MONEA (Message-Oriented NEtworked-robot Architecture) which is an efficient development platform architecture for multifunctional robots \cite{monea}. The architecture embeds a Networked-Whiteboard Model for information sharing framework along with Message passing framework via P2P Virtual Network using Interest-Oriented Module Groups and Software Patterns to reduce complexity risks. It has actually been designed to fulfill three features : embodying the Meta-Architecture for Networked-Robots, supporting Bazaar-Style Development Model, and no need of heavy weight middleware. The MONEA-based middleware has been implemented to develop a dialog robot for exhibition.

On the other hand, there is also another usage of tele-operated mechanism to control and monitor simultaneously several robots over web \cite{amire}. The system provides a comprehensive platform  enabling the users to customize each robot independently.

We follow similar approach in our project to develop a modular and self-independent wireless robot. However, our main objective is completely different, that is to perform more serious tasks requiring telepresence for the reason of safety. For example : direct data retrieval in nuclear reactors, observation apparatus for volcanoes and so on. These purposes lead to completely different requirements. The robot should be able to acquire data in almost real-time basis, and then to process it at the robot's local system as well. Therefore no need for the end-users to install certain softwares in a terminal connecting to the robot. The users just connect their terminals over wireless network and then pointing the browser to the assigned address of robot to display the analyzed results, while at same time it is able to control and monitor the robot's movement, direction etc.

However, in the present paper the discussion is focused on the aspect of microcontroller-based system which is very crucial to realize the a modular system. The system has been applied on a prototype of simple monitoring networked robot without any complex physical movement. First we discuss the concept on our modular networked robot, followed with detailed explanation on the prototype we have developed so far. Finally we summarize the results and discuss future plans.

\section{The concept}

\begin{figure*}[t]
 \centering
 \includegraphics[width=16cm]{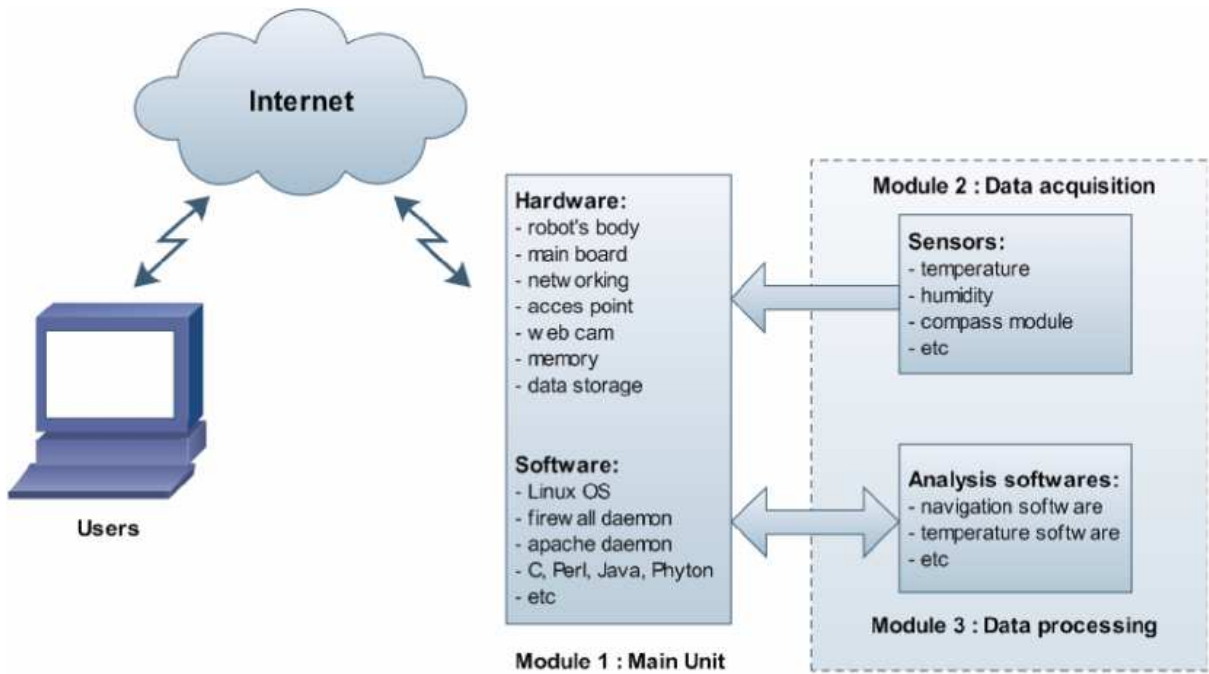}
 \caption{Three modules and its contents constructing the modular robot.}
 \label{fig:modul}
\end{figure*}

Concerning the features mentioned above, we have developed the robot to be as modular as possible to make it more adaptive to users' needs. This concept is depicted in Fig. \ref{fig:modul}, consisting of three modules \cite{lwr2},
\begin{enumerate}
\item Main unit. 
\item Data acquisition module.
\item Data processing module.
\end{enumerate}
So, the system is divided into two independent units : the main unit and the unit set of data acquisition and its processing modules. Let us call the second one briefly as DAPS (Data Acquisition and Processing Systems) throughout the paper. In principle, one might have single main unit with different packages of DAPS, or vice versa according to the needs. This characteristic would enable users to easily replace, for instance the set of censors, actuators and relevant software-based data processing modules with any available packages later on using the same main unit. Inversely, replacing the main unit with more appropriate one to a certain geographical site, but using the same DAPS package.

Just to mention, we are going to open the whole architecture in all modules as an open source. Further this approach could  encourage third parties to develop independently any relevant packages designed for some certain needs. Inversely, another ones who are interested more in hardware developments, might build alternative main unit with different type of mechanics for robot, but with same architecture for the rest to keep its compatibilities with existing DAPS packages. Some considerable packages are, for example, censors of hazard gas combined with software based chemical compound analyzer, vibration censor combined with software based seismograph, and so forth. This is the reason we call it as a generic robot which is adaptable to any DAPS. 

We would like to emphasize that this concept is quite new and a breakthrough in robotic (hardware) technology that is actually inspired by the habit of open-source communities. Also, in contrary most  robots are usually constructed for single particular task. According to this concept, in order to realize high degree of freedom for both users and developers, and to keep full compatibilities in the future developments, let us list common features which should be fulfilled :
\begin{enumerate}
\item All aspects are fully controllable wirelessly over web through TCP/IP protocol.
\item Acquired data are stored and processed at the robot's local system independently from external apparatus.
\item The processed data can be retrieved and analyzed by users also over web such that no need for additional software installation at the user's terminal connecting to the robot.
\item The hardware-based components are replaced as much as possible with the software-based systems in the integrated DAPS packages, even in the main unit.
\end{enumerate}
According to the last point, in particular signal processing and filtering from the censors are performed using customized software rather than hardware as commonly done. This is important again to improve the flexibility and overall cost reduction. Hence, attaching different type of censors would require only different set of DAPS. 

\begin{figure*}[t]
 \centering
 \includegraphics[width=16cm]{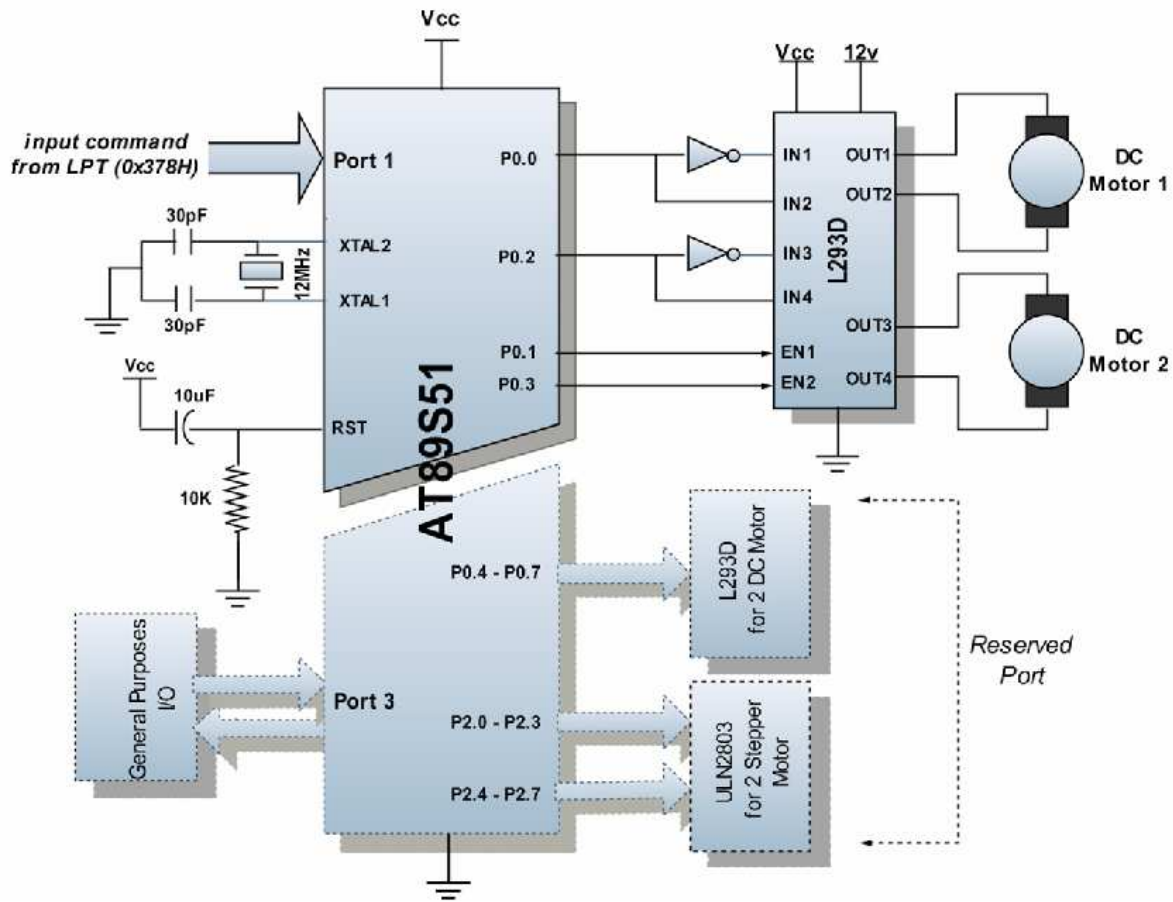}
 \caption{The block diagram for data acquisition, control and monitoring systems in the LNR.}
 \label{fig:diagram}
\end{figure*}

Before going on, let us mention some benefits in deploying this approach :
\begin{itemize}
\item Easy access for end-users regardless the operating system being used. 
\item No need for installing any additional softwares in terminal accessing the robot. 
\item Overall cost reduction, since most components are able to be replaced with software based system.
\item High compatibility due to limited proprietary hardwares and also embedded softwares in the system. \item All software based components are developed using freely available open-source softwares. In our case we use Debian Linux for the operating system, Apache for the web-server and some GNU Public License development languages like GNU C, Java and Python.
\item Highly safe at any untouchable areas of human being, since all aspects of robot are remotely accessible and controllable.
\item Simple calibration since all signals are processed by softwares. This point is crucial regarding the main purpose of direct and continuous retrieval of  physical observables.
\end{itemize}

\section{A prototype : LIPI Networked Robot}

Now we are ready to introduce the LIPI Networked Robot (LNR) that is accessible on the net for public \cite{lwr1}. In developing the prototype, we put the priorities on the DAPS unit rather than the main unit. The reason is the main unit should be simple, at least for our current purpose of autonomous  monitoring works in the limited space.

To be more specific, we have designed a DAPS package to observe hazard gas in a certain area and to measure the surrounding environment. The whole system then consists of,
\begin{itemize}
\item Main unit :\\
	Containing main processor (mini PC etc), storage media (mini hard-disk), access point, power supply, battery and all mechanical components. Of course, it also includes the underlying operating system,  integrated web interface,  hardware control and monitoring systems  and storing system for all  data.
\item Data acquisition module :\\
	A set of censors ($CO$ gas, temperature, humidity, $NO$ gas, smoke) and small camera. The softwares in this  module are also responsible for filtering the raw signal from censors. 
\item Data processing module :\\
	It covers all add-on softwares to process, store and analyze the acquired data. 
\end{itemize}

\begin{figure}[t]
 \centering
 \includegraphics[width=9cm, height=6cm]{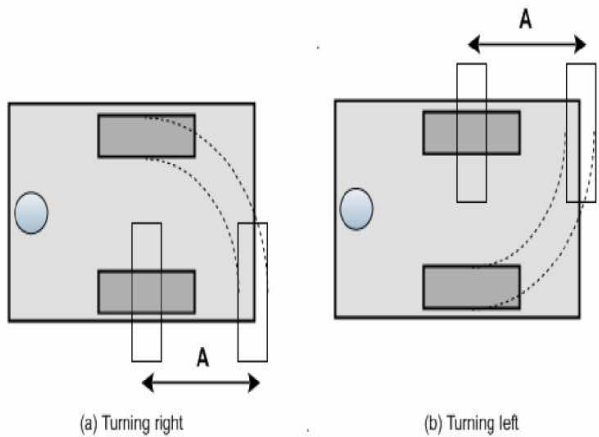}
 \caption{The moving characteristics of the wheel control.}
 \label{fig:roda}
\end{figure}

The DAPS unit is depicted diagrammatically in Fig. \ref{fig:diagram}. In this design, we use the  parallel port for both reading the signals from and sending some commands to microcontroller. As a result, the microcontroller decodes each command into an associated task such as driving the actuators (in the case of LNR are two DC motors), and also retrieving  raw signals from attached censors. Clearly, the microcontroller plays important roles in this design. As depicted in Fig. \ref{fig:diagram}, the main controller is AT89S51 microcontroller. The microcontroller in this design is responsible not only for receiving all incoming command sent from main module but also for controlling all the actuators such as DC and stepper motors. In order to make the robot to be as modular as possible, more I/O's are needed. In the present case, we have developed an integrated  interface for totally 4 DC motors, 2 stepper motors and one byte additional general purposes I/O port. 

In principle, we should provide as many channels as possible to attach various censors and motors. All of them are later combined and controlled to realize a particular actions along with the purposes. More importantly, all combinations are maintained by certain software-based package in the main unit and DAPS that should be developed in accordance with the purposes.

In this paper, however we are not going to discuss the microcontrollers in the DAPS unit since the detail requires the DSP rather than the microcontroller itself \cite{lwr2}. Instead, we describe the way to deal with the wheels as actuators and the associated dead reckoning to track the current position easily using the microcontroller based hardwares.

\subsection{Actuators}

Now let us discuss the actuators in LNR to enable movements in any directions. DC and stepper motors are  widely used in robotics due to its simplicity in interfacing and low cost as well. Both can be used for continuous rotary motion or precise angular displacement. However, LNR makes use of only 2 DC motors concerning the energy consumption  to save the energy resource which in the case of LNR is only a small battery. 

Further we deploy the $H-$bridge method to realize forward and backward movements \cite{hbridge}. Then  combining two DC motors attached to two independent wheels would enable multi-directional movements. The $H-$bridge circuits principally consist of four switches with specific configurations of switching to control the electric current to the DC motor.  Switching one input ($X$) to low and another input ($Y$) to high state yield the upper right and lower left closed and the other two remain open, and so forth. As seen  in Fig. \ref{fig:diagram}, we use IC L293D as motor driver to simplify the circuit  design. L293D consist of dual H-Bridge motor driver and diodes inside for protecting the circuit from EMF outputs. Basically to control each DC motor we need three inputs of L293D which is controlled by AT89S51 microcontroller subsequently. However, using inverter logic IC such as 74LS04 the number of pins controller can be reduced, that is only pins 0 and 1 of Port 0 are necessary. The table below shows how to control DC motor using L293D as a dual H-Bridge motor drivers. Therefore, using 4 combinations of values taken by IN1 and IN2 is more than enough to define specific commands to rotate a motor clockwise, anti-clockwise and stop.

The block diagram of module for controlling the wheels is given in figure of Fig. \ref{fig:diagram}. In this design, we use AT89S51 microcontroller from Atmel as wheel controller. This type of microcontroller is efficient to control the DC motors and has good cost performance since it has no unnecessary feature inside it such as on-chip ADC, PWM generator, etc. To build the DC motor controller, besides the $H-$bridge circuits, we also need a current driver circuits to amplify the electric current to accelerate the motors directly. Using only one parallel port in microcontroller such as PA,PB,PC or PD, we can control two motors simultaneously  \cite{braunl}\cite{mccomb}. 

For the wheel structure, we have implemented the simplest mechanical design in LNR to avoid unnecessary complexities since our focus is on the DAPS unit rather than the main unit as mentioned previously. The mechanical components consists of three wheels, one is freely rotateable while the other two are fixed and driven by two independent DC motors. Using these motors we are able to control the robot movement in four directions such as turning left and right, moving forward and backward by controlling each DC motors independently as shown in the right figure of Fig. \ref{fig:roda}. Turning left (right) is realized by stopping the left (right) motor and starting the right (left) motor respectively. Moving forward is simply done by starting both motors simultaneously.

\subsection{Dead reckoning}

\begin{figure*}[t]
 \centering
 \includegraphics[width=9cm]{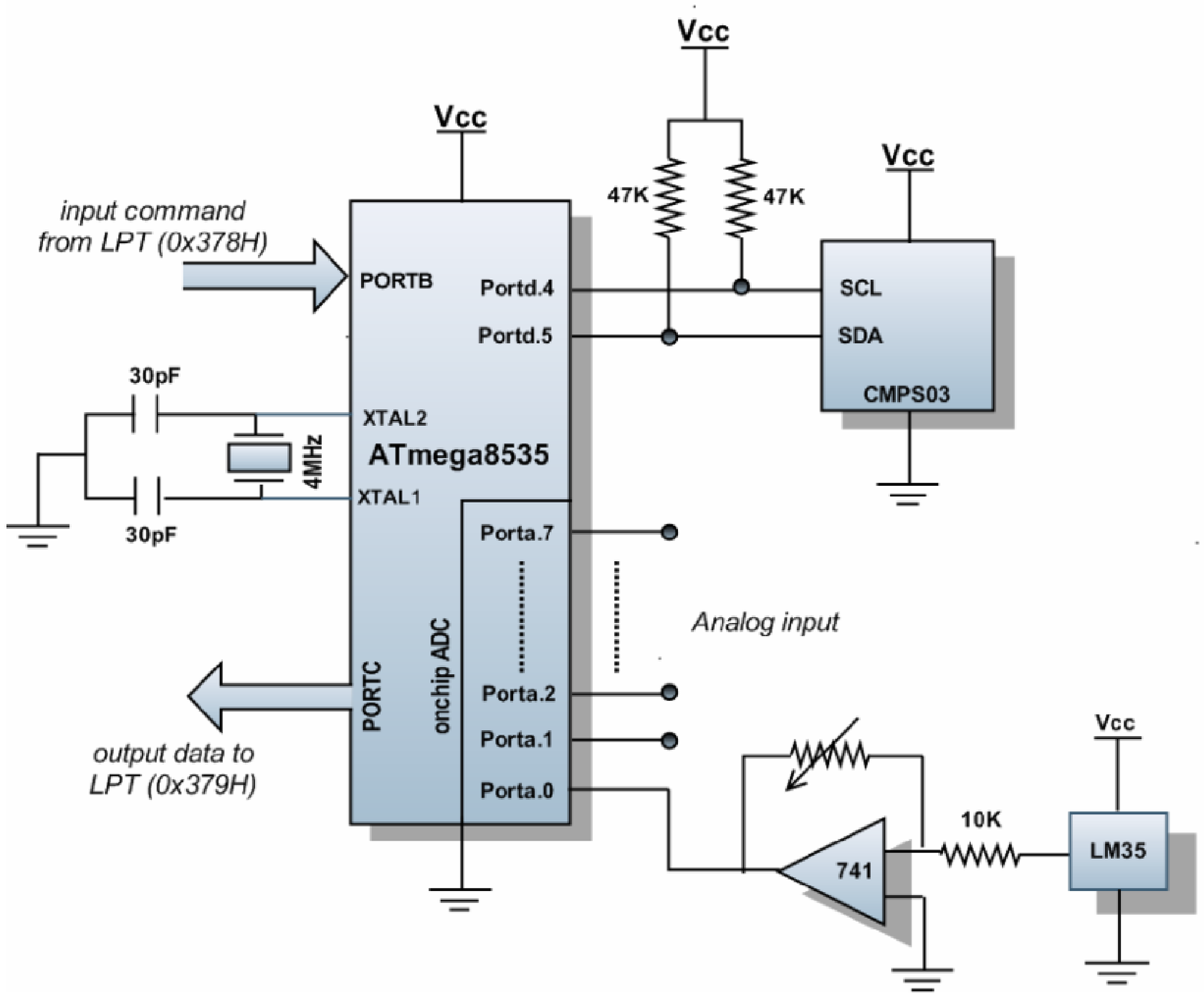}
 \hfil
 \includegraphics[width=9cm]{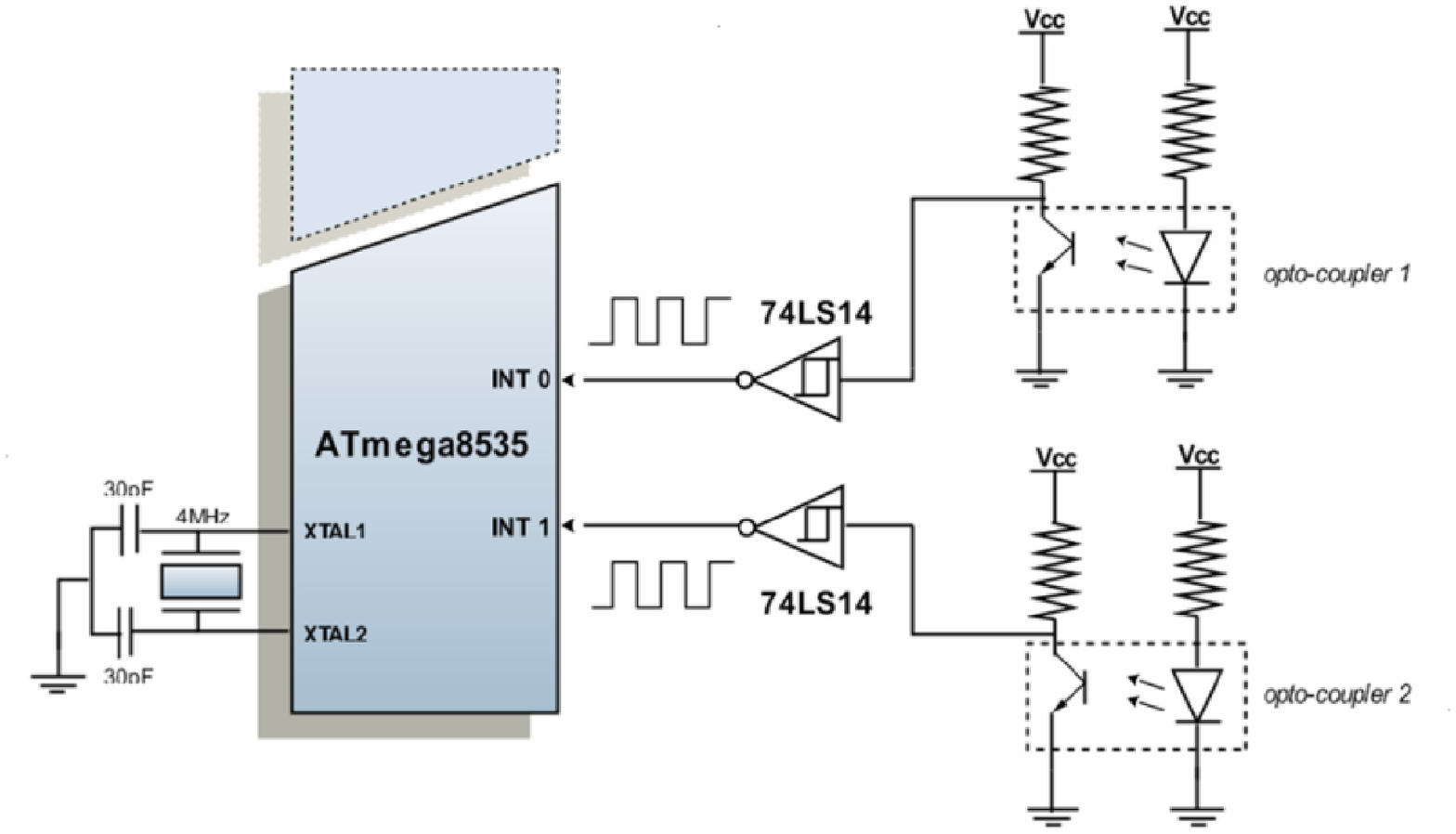}
 \caption{Interfacing ATmega8535 with CMPS03 (left), and the wheel rotation counter (right).}
 \label{fig:kompas}
\end{figure*}

Next issue is how to recognize the current position at almost real-time basis. Instead of using the images captured by camera which would require complicated and resource wasting image processing, we make use of the compass censor and the wheel rotation counters. Again, we remind that in LNR the main processor should be prioritized to retrieve, store and analyze the data.  Although a small camera is attached on the robot, it is only intended to get the actual visualization around the robot.  Therefore, the problem is turned into how to track the paths and recognizing the current position in a sightless condition.

The compass censor provides information of the actual angle against the North-South pole, while counting each wheel rotation yields the point-to-point distance. This mechanism is later on visualized on the web in real-time basis as a virtual compass and the footprints of wheels from one point to another. In LNR we make use of a  digital compass module CMPS03 \cite{coe}. This compass module has been specifically designed for a usage in robots as add-on navigation. It generates a unique number in byte from 0 to 255 to represent the angles within $0^\mathrm{o}\sim360^\mathrm{o}$. It is also embedded with two   Philips KMZ51 magnetic field censors which are sensitive enough to detect the earth magnetic field. The internal microcontroller converts the signals from magnetic field censor into serial I2C data format. We have deployed a typical hardware configuration for interfacing CMPS03 compass module using I2C interface as depicted in Fig. \ref{fig:kompas}. The figure describes a system for a single motor. In principle the same interface should be deployed for each channel of actuator provided in the main unit to enable independent counter for each motor.

Using the compass module, the angles information is retrieved through I2C communication protocol by sending a start bit, the module address (0XC0) with the read/write bit low and the register number we wish to read. This is followed by a repeated start and the module address again with the read/write bit high (0XC1). This procedure allows us to read one or two bytes for 8-bit or 16-bit registers respectively. This type of communication algorithm can be seen in the documentation of CMPS03 \cite{coe}. However, this module also provides alternative data format, that is it generates the serial  data output in PWM (pulse width modulation) mode. The PWM signal is a pulse width modulated signal which varies from 1 ms (equivalent to $0^\mathrm{o}$) to $36.99$ ms (equivalent to $359.9^\mathrm{o}$).

In order to recognize a real path length, we simply count the wheels rotation whenever the wheels start moving and stopping. The wheel rotation is measured by using opt coupler. The opt coupler detects a black and white colors on the wheel and produces digital signals (1 and 0). To sharpen the transition level in digital signal produced by opt coupler, we use 74LS14 a Schmitz trigger IC. In this design, the external interrupt INT0 and INT1 on PORTD bit 2 and bit 3 can serve as an external interrupt source to the ATmega8535 microcontroller as can be seen in Fig. \ref{fig:kompas}. This outputs pulse can be directly counted by microcontroller through external interrupt INT0 and INT1 pin. In software side, the interrupt method is used rather than polling methods. In LNR each wheel is divided into eight areas which means a complete rotation occurs after eight counts, however we can divide it as much as possible to improve the accuracy.

All information of the angles and the path lengths are recorded path by path. Finally we borrow the dead reckoning method which is very powerful to calculate the position relying on a previously determined position \cite{dr}. To implement such algorithms on a robot, we have constructed a mathematical prescription to calculate the geometrical distances of each path obtained from each wheels counter, and its relative angles obtained from the compass censor. The final formula is embedded into the software belonging to that main unit. The formula provides an exact calculation for each intermediate path, the total running-distance over the paths and the real distance between the initial and end points. Nevertheless, due to the limited space the detail algorithm and formulae will be given elsewhere.

\section{Summary and discussion}

We have introduced a concept of modular networked robot and its microcontroller-based system. A detail design for the actuators are presented. As a typical implementation of the concept, we have developed the LNR. The robot has some unique features as mentioned in the preceding sections. The main advantage is its flexibility to various purposes. We argue that regarding its main objective as a monitoring apparatus, LNR is quite efficient and has good total cost-performance due to its modularity and dominant software based solutions. Actually the current LNR is expandable to be attached up to six independent censors and six independent actuators.

However, the system has complicated aspects and still requires further developments  :
\begin{itemize}
\item More examples of DAPS packages fit certain purposes. 
\item More complicated robot's actuators and the relevant algorithms for its software-based control and monitoring systems.
\item Further development of main unit and its mechanical components to enable more advanced and smooth movements. Rather than full hardware-based approach, this will be done by utilizing as much as possible integrated software-based algorithms.
\item Automatic and software based calibration systems to keep the accuracy of actuators. For instance  in the current LNR, synchronizing the rotation speed of left and right wheels, reseting the initial angle of camera and so on.
\item Lastly, we would like to announce that the architecture and all related softwares of LNR will be open for public under GNU Public License once we consider the system is ready for further development by open-source communities around the world \cite{opennr}.
\end{itemize}

\section*{Acknowledgment}

The work is financially supported by the Riset Kompetitif LIPI in fiscal year 2008.

\end{document}